\documentclass{article}

\usepackage{PRIMEarxiv}

\usepackage[utf8]{inputenc} 
\usepackage[T1]{fontenc}    
\usepackage{hyperref}       
\usepackage{url}            
\usepackage{booktabs}       
\usepackage{amsfonts}       
\usepackage{nicefrac}       
\usepackage{microtype}      
\usepackage{lipsum}
\usepackage{fancyhdr}       
\usepackage{graphicx}       
\graphicspath{{media/}}     
\usepackage{amsmath}
\usepackage{algorithmic}
\usepackage{algorithm}
\usepackage{array}
\usepackage{booktabs}
\usepackage{textcomp}
\usepackage{stfloats}
\usepackage{verbatim}
\usepackage{graphicx}
\usepackage{indentfirst}\usepackage{indentfirst}
\usepackage{balance}
\usepackage[justification=centering]{caption}

\usepackage{ragged2e} 
\usepackage{booktabs,makecell, multirow, tabularx}
\title{Intelligent: Optimizing High-Level Synthesis Design Space Exploration with Graph Neural Networks and Large Language Models

}

\author{
  Lei Xu, Shanshan Wang \\
  Department of Computer Science \\
  Shantou University \\
  Shantou, China\\
  \texttt{\{24lxu, sswang\}@stu.edu.cn} \\
   \And
  Emmanuel Casseau \\
  IRISA \\
  Unviersity of Rennes 1 \\
  Rennes, France\\
  \texttt{emmanuel.casseau@irisa.fr} \\
  \And
   Chenglong Xiao \thanks{\textit{\underline{Corresponding Author}}: 
\textbf{Chenglong Xiao.}}\\
  Department of Computer Science \\
  Shantou University \\
  Shantou, China\\
  \texttt{chlxiao@stu.edu.cn}  \\
}

\begin{document}

\maketitle

        \begin{abstract}
		High-Level Synthesis (HLS) Design Space Exploration (DSE) is essential for generating hardware designs that balance performance, power, and area (PPA). To optimize this process, existing works often employs message-passing neural networks (MPNNs) to predict quality of results (QoR). These predictors serve as evaluators in the DSE process, effectively bypassing the time-consuming estimations traditionally required by HLS tools. However, existing models based on MPNNs struggle with over-smoothing and limited expressiveness. Additionally, while meta-heuristic algorithms are widely used in DSE, they typically require extensive domain-specific knowledge to design operators and time-consuming tuning. To address these limitations, we propose ECoGNNs-LLMMHs, a framework that integrates graph neural networks with task-adaptive message passing and large language model-enhanced meta-heuristic algorithms. Compared with state-of-the-art works, ECoGNN exhibits lower prediction error in the post-HLS prediction task, with the error reduced by 57.27\%. For post-implementation prediction tasks, ECoGNN demonstrates the lowest prediction errors, with average reductions of 17.6\% for flip-flop (FF) usage, 33.7\% for critical path (CP) delay, 26.3\% for power consumption, 38.3\% for digital signal processor (DSP) utilization, and 40.8\% for BRAM usage. LLMMH variants can generate superior Pareto fronts compared to meta-heuristic algorithms in terms of average distance from the reference set (ADRS) with average improvements of 87.47\%, respectively. Compared with the SOTA DSE approaches GNN-DSE and IRONMAN-PRO, LLMMH can reduce the ADRS by 68.17\% and 63.07\% respectively. 
    \end{abstract}

	\section{Introduction} 
    High-Level Synthesis (HLS) has emerged as a modern alternative to traditional RTL-based VLSI design. HLS enables users to leverage high-level programming languages, such as C/C++, for hardware design tasks, offering synthesis options to fine-tune results. Proven in practice, HLS provides numerous advantages in hardware design and has become the mainstream approach, particularly for FPGA implementations. To enhance HLS effectiveness, most Electronic Design Automation (EDA) tools offer a variety of synthesis options, called "knobs", allowing users to control the synthesis outcomes. In commercial HLS tools, synthesis directives are typically added to the source code as comments or pragmas. By evaluating various directive combinations, HLS produces a wide range of hardware implementations, facilitating the exploration of trade-offs between conflicting objectives such as performance (latency) and cost (area, power).
    
	\begin{figure*}[t]
		\centering
		\includegraphics[width=0.75\linewidth]{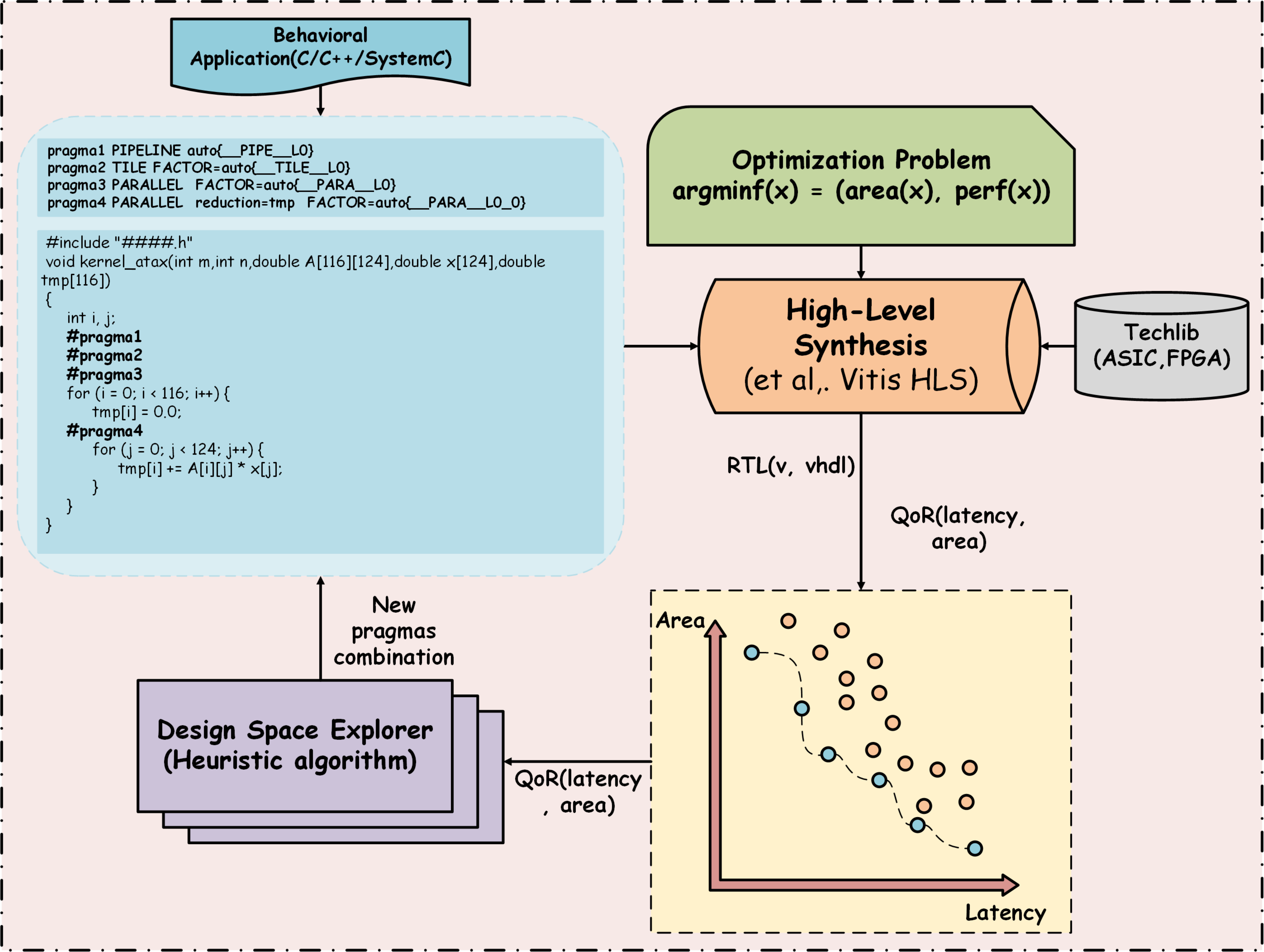}
		\caption{A comprehensive overview of the HLS DSE process uses the \textit{atax} application from the PolyBench benchmark suite. The HLS tools can be replaced by predictive models.}
		\label{HLSDSE1}
         \vspace{-5mm}
	\end{figure*}
	
	Fig. \ref{HLSDSE1} presents an overview of high-level synthesis design space exploration. The workflow begins with C/C++ behavioral descriptions augmented with pragma directives and parameterized optimization targets. These quality of results (QoR) metrics (e.g., latency, resource utilization) of the specifications can be obtained using HLS tools or machine learning based models. A design space exploration engine then selects novel pragma combinations through multi-objective optimization, generating updated behavioral descriptions for subsequent iterations. This cyclic refinement process continues until the convergence to a Pareto-optimal configuration set that optimally balances competing design constraints. Therefore, accurate QoR prediction of each design and efficient design space exploration are the most crucial tasks in HLS DSE. Regarding the two key tasks involved in HLS DSE, we identify the following critical issues:
	
	\textit{Problem 1 (High-accuracy QoR prediction)}: Current QoR prediction methodologies predominantly employ graph representation learning, where pragma-annotated source code is transformed into control-data flow graph (CDFG), followed by sophisticated GNN architectures extracting topological features for predictive modeling. GNNs based on the message-passing mechanism, commonly termed MPNNs, operate by iteratively updating node features through neighborhood information aggregation. Although these MPNNs demonstrate strong accuracy in conventional classification and regression tasks, their performance in QoR prediction tasks remains limited, often failing to surpass the discriminative power of the 1-Weisfeiler-Lehman (1-WL) algorithm \cite{Finkelshtein}. This limitation comes from their restricted expressiveness, leading to sub-optimal performance in complex prediction tasks \cite{Xu}. In addition, conventional MPNNs suffer from inherent long-range dependency issues and over-smoothing phenomena \cite{Li}, severely constraining GNN models' accuracy and generalization capabilities in QoR prediction tasks. The problems existing in the above-mentioned MPNNs all seriously affect their QoR prediction accuracy in post-HLS and post-implementation. A critical observation reveals that existing GNN models seldom excel in both post-HLS and post-implementation QoR prediction tasks, inherently linked to the architectural limitations of MPNNs.
		
	\textit{Problem 2 (Multi-objective design exploration)}: DSE methods systematically explore the design space to identify Pareto-optimal configurations that achieve optimal trade-offs between competing objectives such as performance and cost. In DSE, meta-heuristic algorithms (non-dominated sorting genetic algorithm II (NSGA-II) \cite{Schafer1}, simulated annealing (SA) \cite{Schafer3}, and ant colony optimization (ACO) \cite{Schafer2}) have profound applications and have achieved remarkable success. One critical issue lies in the inherent challenge for meta-heuristic algorithms to identify superior design configurations within vast design spaces while exploring a limited number of design candidates. While traditional meta-heuristics demonstrate satisfactory performance on benchmarks with constrained design spaces, their effectiveness fundamentally degrades when handling exploration tasks encompassing millions or even hundreds of millions of configurations. Under such scale constraints, the optimization objective shifts from identifying the absolute optimal configuration to discovering acceptably competitive solutions within strictly bounded exploration iterations and time budgets. Conventional approaches relying solely on parameter configurations and stochastic operators prove inadequate for achieving viable design solutions through sparse exploration of the combinatorial space. Another critical challenge stems from the parameter sensitivity inherent in meta-heuristic algorithms. Table \ref{ap} presents the algorithm parameters employed by NSGA-II, ACO, and SA. The performance of these algorithms depends heavily on parameter settings. For instance, the crossover and mutation rates in NSGA-II should be adjusted according to the design space size. Generally, for larger benchmarks, the crossover and mutation rates should be set slightly higher to avoid falling into local optima. However, manual parameter tuning for distinct benchmarks remains impractical, constituting a fundamental limitation of meta-heuristic algorithms.
	
	\begin{table}
		\renewcommand{\arraystretch}{1.2}
		\centering
		\tiny
		\caption{Key Parameters and Descriptions for NSGA-II, ACO, and SA}
		\label{ap}
		\begin{tabular}{c|c|c}
			\toprule
			{\bf{Algorithms}}&{\bf{Parameters}}&{\bf{Description}}\\
			\midrule
			\multirow{2}*{NSGA-II} & Crossover Probability & The probability that two parent solutions will undergo crossover to produce offspring.	\\
			\cline{2-3}
			~ & Mutation Probability & The probability that a variable in a solution will be mutated.\\
			\hline
			\multirow{4}*{ACO} & Pheromone Factor ($\alpha$) &  The weight of the pheromone trail in the decision rule.\\
			\cline{2-3}
			~ & Heuristic Factor ($\beta$) & To balance exploration and exploitation in the control algorithm.\\
			\cline{2-3}
			~ & Evaporation Rate ($\rho$) & The rate at which pheromone evaporates each iteration.\\
			\cline{2-3}
			~ & Pheromone Constant ($Q$) & A constant related to the total amount of pheromone an ant deposits.\\
			\hline
			\multirow{2}*{SA} & Perturbation Probability& The global probability that any given parameter in the configuration vector will be selected for a change. \\
			\cline{2-3}
			~ & Neighborhood Radius & The maximum permissible distance between the current solution and a newly generated candidate solution \\
			\bottomrule
		\end{tabular}
		\vspace{-6mm}
	\end{table}	
	
	To address the aforementioned issues, we propose ECoGNNs-LLMMHs, a unified framework that synergizes  task-adaptive dynamic graph neural networks with large language model (LLM)-driven meta-heuristic algorithms. Our key contributions are as follows:

	\begin{itemize}
		\item [1)] We adopt a novel graph neural network architecture (ECoGNNs) with an adaptive message-passing mechanism as our core prediction model for accurate QoR prediction. Unlike traditional GNNs with fixed message-passing rules, ECoGNNs dynamically reconstruct graph topologies based on task objectives, enabling targeted feature propagation and mitigating over-smoothing. The adapted ECoGNNs can be tuned to predict both post-HLS and post-implementation QoR metrics with the highest accuracy compared to those of state-of-the-art (SOTA) works.
        \item [2)] We propose an LLM-driven meta-heuristic (LLMMH) framework for efficient DSE. The evaluated meta-heuristic algorithms include NSGA, SA, and ACO. By leveraging LLMs’ in-context learning (ICL) capabilities with carefully engineered prompts, LLMMH significantly reduces domain expertise dependency and achieves significant improvement in Pareto front quality compared to baseline algorithms. Moreover, the proposed LLMMH framework can be adapted to many other meta-heuristic algorithms for HLS DSE.
        \item [3)] To the best of our knowledge, the proposed LLMMH is the first framework that integrates an LLM with meta-heuristic algorithms for HLS DSE. This approach represents a significant shift from prior methods in HLS DSE. Notably, it has demonstrated a promising ability to enhance the quality of DSE outcomes. We hope that our findings will inspire further exploration of LLM-based techniques to tackle challenges in HLS DSE.
    \end{itemize} 			
	
	\section{Related work}
	{\bf{QoR prediction.}} Recent advances in QoR prediction have been driven by machine learning and GNNs. Foundational approaches employ learnable parameters and loss minimization to predict optimal configurations, with notable contributions including systematic regression model analysis for HLS feature selection \cite{Dai} and the accelerator evaluation framework MPSeeker \cite{Zhong}. Makrani et al.  \cite{Makrani} employ Minerva, an automated hardware optimization tool that determines near-optimal tool configurations through static timing analysis and heuristic algorithms, optimizing either peak throughput or throughput-to-area ratio. The Pyramid framework leverages this database to train an ensemble ML model that maps HLS-reported features to Minerva's optimization outcomes.Building on these foundations, GNN-based methods have emerged as the dominant paradigm, typically converting source code into intermediate representation (IR) graphs for feature extraction exemplified by hybrid control-data flow graphs \cite{Ferretti} and hierarchy-aware GNNs \cite{Gao}. Recent studies demonstrate progressive advancements in GNN-based hardware design optimization. Wu et al. \cite{Wu1} develop performance modeling frameworks using graph neural networks to represent C/C++ programs as computational graphs. Extending this paradigm, Zhao et al. \cite{zhao} establish GNNHLS as an HLS-specific benchmark with six GNN models across four topological datasets. While Lin et al. \cite{Lin} propose power-aware edge-centric GNNs with dynamic power modeling through edge neighborhood aggregation, Kuang et al. \cite{Kuang} introduce hierarchical HGP architectures for post-implementation PPA prediction. Complementing these approaches, Sohrabizadeh et al. \cite{Sohrabizadeh1} present HARP's hierarchical graph representation with auxiliary nodes for pragma-optimized HLS designs.
	
	{\bf{Design space exploration.}} DSE is fundamentally formulated as a multi-objective optimization problem (MOOP), where pre-trained prediction models can serve as evaluators to guide heuristic search algorithms toward Pareto-optimal solutions. Wang et al. \cite{Wang} automate meta-heuristic parameter configuration through behavioral analysis of C-based descriptions, while Goswami et al.  \cite{Goswami} demonstrate that GBRT-based models achieve comparable accuracy to exhaustive logic synthesis. To enable a precise identification of Pareto-optimal designs, Hong et al. \cite{Hong} develop contrastive learning to predict the dominance relationships between designs. A reinforcement learning-driven DSE approach is presented in the work \cite{Wu2}. Yao et al.  \cite{Yao} propose a decomposition based DSE approach to minimize synthesis iterations. To accelerate ASIC DSE,  Rashid et al. \cite{Rashid} introduce FPGA-targeted transfer learning and a dedicated multithreaded parallel HLS design space explorer.
	
	In summary, SOTA prediction methodologies in HLS depend primarily on GNNs, leveraging their inference capabilities to achieve high-precision predictions on unseen datasets. However, inherent limitations of MPNNs, such as long-range dependency failures and over-smoothing phenomena, critically degrade prediction accuracy, partially explaining the persistent performance gaps in HLS QoR estimation. Meanwhile, current improvements in graph representation methods can provide prediction models with more information contained in the source code. For example, HARP \cite{Sohrabizadeh1} incorporates pragma information into graph representations and utilizes encoders for joint feature extraction of both this information and graph structural data. However, due to the insufficient expressive ability of the convolutional ($conv$) layers used in HARP, the actual improvement it brings is relatively limited compared with that of GNN-DSE \cite{Sohrabizadeh}. The fundamental constraint remains the restricted feature extraction capability of graph encoders, indicating insufficient expressive power. In addition, due to the inconsistent sources of data labels, some derived from HLS results and others from post-implementation results, previous works have not simultaneously verified the model's models' performance on these two types of datasets, thus potentially suffering from the limitation of narrow application scenarios. The recent GNN architecture CoGNNs \cite{Finkelshtein} demonstrate adaptive capability with dynamic message-passing mechanism, which is particularly beneficial for HLS prediction tasks. In this work, we adopt Enhanced-CoGNNs (ECoGNNs) for predicting post-HLS and post-implementation metrics. 
	
	Meta-heuristics have been the predominant approaches and have achieved significant progress in solving HLS DSE over the past decades, and they continue to undergo diverse and flourishing development. The main problems of the traditional meta-heuristic algorithms lie in the difficulty of the designed random operators in escaping the local optimum and the numerous difficulties in parameter tuning. Inspired by the works proposed in \cite{Liu,Meyerson}, we propose an LLM-driven meta-heuristic framework for HLS DSE, which takes advantage of LLMs' ICL capabilities to interpret the functional roles of various pragmas and their impacts on the final routing outcomes.
	
	\section{Preliminaries}
    	
	\subsection{Problem Formulation}
	HLS DSE can be formulated as a MOOP, where the primary objective is to identify a set of Pareto-optimal configurations that simultaneously minimize latency and resource utilization. Given a behavioral description and optional synthesis directives $[pragma_{1}, pragma_{2}, ...,pragma_{n}]$, the design space $DS$ can be formulated as the Cartesian product of the combinations of pragmas. The definition of Pareto frontier $P$ is as follows \cite{Schafer}:
	\begin{equation}
		A(d_P) \leq A(d_i)\  and\  L(d_P) \leq L(d_i)
		\label{4}
	\end{equation}
	where $d_{P}\in P$, $d_{i} \in DS$ and $P \subseteq DS$. The functions A(·) and L(·) represent the resource utilization and the latency of $d_P$ and $d_i$, respectively. A design configuration $d \in P$ is considered Pareto-optimal if no other configuration in the search space dominates it, that is, no alternative design simultaneously achieves both lower resource utilization and reduced latency. Therefore, our objective is to efficiently identify the Pareto-optimal set without resorting to exhaustive searches across the entire configuration space or invoking computationally expensive HLS tools.
	   	
	\subsection{Meta-heuristic Algorithms for DSE}
	In DSE, NSGA-II \cite{Schafer1} is widely adopted as the optimization backbone, where each design configuration is encoded as a gene and the complete set forms a chromosome $C_r$. The exploration begins with an initial population of parent configurations, which undergo crossover and mutation operations guided by predefined probabilities to generate offspring by modifying pragma values (e.g., loop unrolling factors, array partitioning schemes). SA \cite{Schafer3} simulates the natural annealing process by dynamically calculating the probability of accepting either inferior solutions or neighboring configurations during each iteration, with acceptance criteria governed by current temperature parameters. ACO \cite{Schafer2} emulates ant colony behavioral mechanisms by constructing pheromone matrices for potential pragma values. During each iteration, it probabilistically selects pragma configurations through pheromone-guided stochastic sampling using sophisticated probabilistic models, thereby generating solutions with enhanced quality potential.
	
	\subsection{In-context Learning} \label{icl}
    In-context learning (ICL) \cite{min2022} enables large language models to accurately perform downstream tasks by processing natural language prompts containing demonstrations and instructions, requiring no fine-tuning. The in-context learning capability of LLMs stems from massive corpora and complex training processes, endowing them with robust natural language understanding capabilities. The in-context learning capability of LLMs stems from massive corpora and complex training processes, endowing them with the ability to understand and infer underlying concepts. Specifically, after a pre-trained LLM receives an in-context prompt $C_p$ that contains $K$ example pairs ($c_{j}$, $r_{j}$), current knowledge $Know$ and a test input $c_{test}$, it selects the output sequence with the highest conditional probability as the final answer $\hat{r}_{test}$, and this process can be expressed as follows:
    \begin{equation}
    	 \hat{r}_{\text{test}} = \arg\max_{r} P(r \mid c_{K}, (c_1, r_1), \ldots, (c_K, r_K), c_{\text{test}}; \theta)
    \end{equation}
    
	where $P$ refers to the probability distribution that LLMs use to generate $r_{test}$ based on the knowledge acquired from previous text (i.e., weights and parameters $\theta$), combined with the example input-output ($c_{j}$, $r_{j}$) pairs and knowledge $Know$ contained in the in-context prompt. We propose leveraging the in-context learning capability of LLMs to enhance traditional metaheuristic algorithms for more intelligent design space exploration. 
    
	\begin{figure*}[t]
		\centering
		\includegraphics[width=\linewidth]{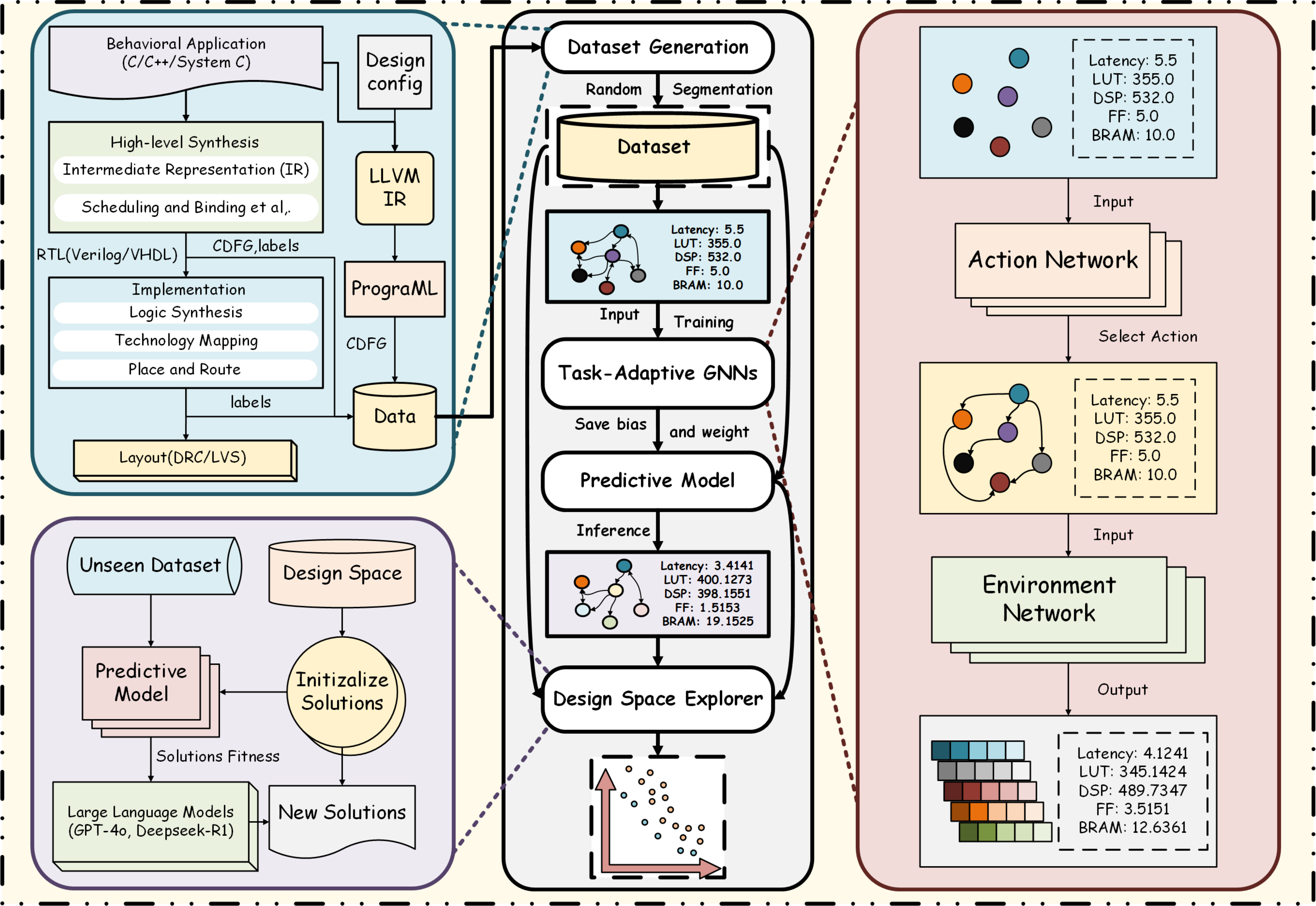}
		\caption{The ECoGNNs-LLMMH framework integrates data generation (combining HLS reports and implementation reports), predictive model training via task-adaptive graph neural networks, and DSE using an LLM-enhanced evolutionary algorithm. The ECoGNNs can be tuned to predict post-HLS QoR metrics and post-implementation QoR metrics.}
		\label{ECoGNNs-LLMEA framework}
         \vspace{-5mm}
	\end{figure*}
	
	\section{Methodology}
	Fig. \ref{ECoGNNs-LLMEA framework} provides an overview of the proposed ECoGNNs-LLMMH. The proposed framework integrates four key stages: dataset generation, model training, inference, and DSE. During dataset generation, labels are extracted from both HLS reports (e.g., latency, resource estimates) and implementation reports (e.g., post-place-and-route resource usage, timing closure) to enhance prediction accuracy. For model training, a subset of the dataset is used to train ECoGNNs composed of an environment network (global feature extraction), an action network (dynamic topology adjustment), and a multi-layer perceptron (QoR metric mapping) which iteratively improves prediction accuracy through dynamic feature extraction. In the inference stage, the trained ECoGNNs model predicts QoR metrics for unseen configurations, enabling rapid evaluation without invocations of the HLS tool. Finally, the DSE stage employs modified meta-heuristic algorithms enhanced by LLMs to identify near-Pareto-optimal solutions.
	
	\subsection{Dataset Generation}\label{III-1}
	The C/C++ source code is first converted to an intermediate representation (IR) using LLVM, followed by the generation of a CDFG through ProGraML \cite{Chris}, which incorporates HLS directives for enhanced semantic representation. Concurrently, following the GNN-DSE \cite{Sohrabizadeh}, the pragma inserter constructs icmp nodes to augment the CDFG with pragma-specific information. The CDFG generated through this compilation process serves as the foundational input to the prediction model. To validate the performance of our proposed prediction model under different scenarios, we train our prediction model on datasets from HGBO-DSE \cite{Kuang} and GNN-DSE \cite{Sohrabizadeh}. For the GNN-DSE dataset, the labels for each benchmark are extracted from the HLS reports, and the CDFGs are generated by the compilation front-end. For the HGBO-DSE dataset, the labels are derived from the implementation reports, and the CDFGs come from the HLS process. The two aforementioned datasets differ not only in their label sources but also in the benchmarks used for their construction. The details of node features in the GNN-DSE dataset are shown in Table \ref{t1}. Categorical features such as node type and instruction type are represented using one-hot encoding, while preprocessed numerical features are directly utilized. The node features of the HGBO dataset are described in its corresponding paper.

	\begin{figure*}[t]
		\centering
		\includegraphics[width=0.8\linewidth]{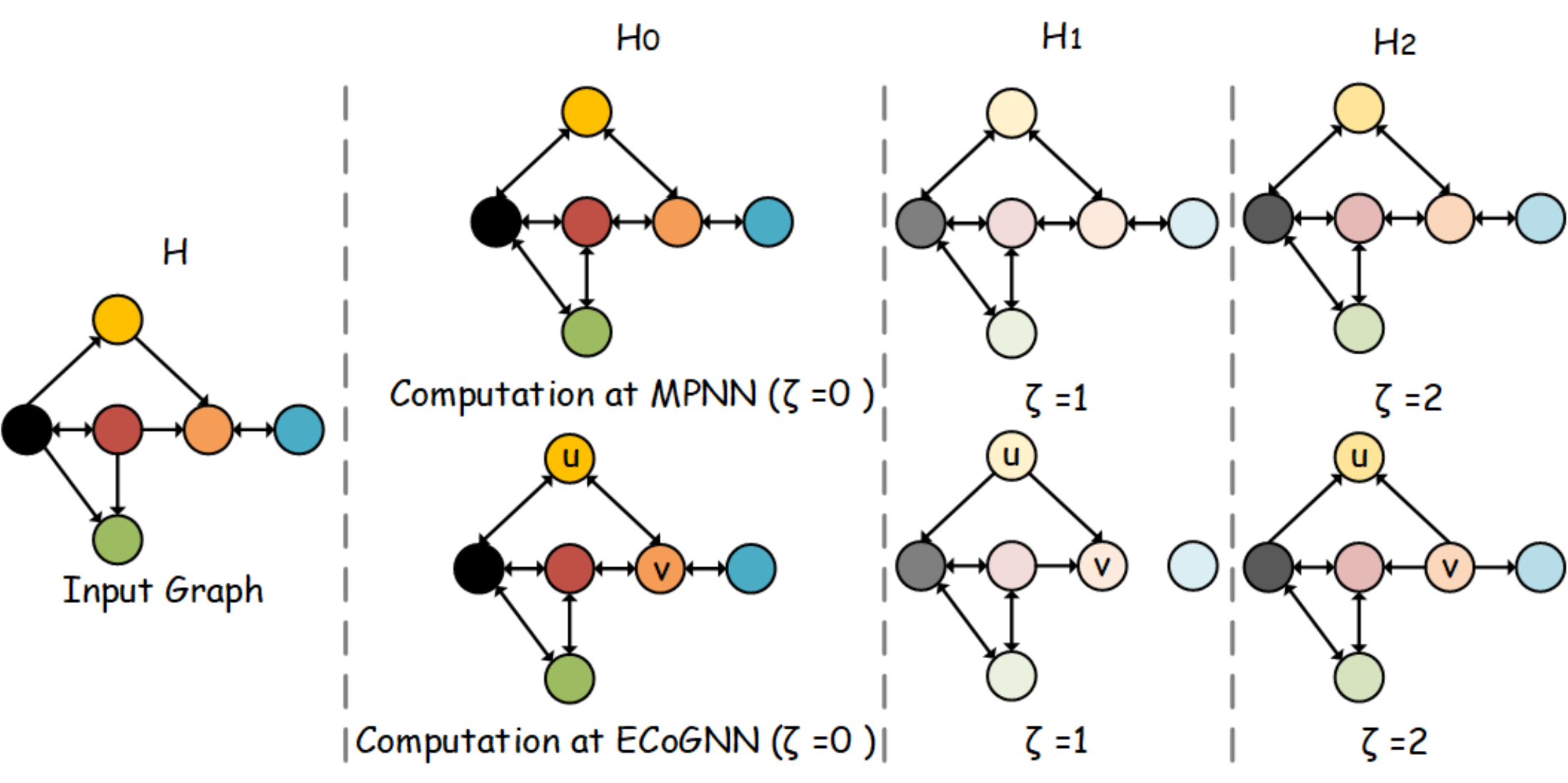}
		\caption{The input graph $H$ and the computational graphs $H_0$, $H_1$, $H_2$ in MPNN and ECoGNN, where the computational graphs serve as an abstract representation of the message-passing process, with arrow directions indicating the flow of information.}
		\label{nc}
		\vspace{-3mm}
	\end{figure*}
	
	\begin{table}[t]
		\small
		\setlength{\tabcolsep}{0.1pt}
		\centering
		\caption{Node Features of Graph Data.}
		\begin{tabular}{c c c}
			\toprule
			\bf{Features} & \bf{Description} & \bf{Values type} \\
			\midrule
			\textit{node type} & index of node type & long \\ 
			\textit{instruction type} & index of function & string \\
			\textit{function type} & index of function & long \\
			\textit{block type} & index of function & long \\
			\textit{Latency} & amount of clock cycles in the node & int \\
			\textit{LUT} & amount of LUT used in the node & int \\
			\textit{DSP} & amount of DSP in the node & int \\
			\textit{FF} & amount of FF in the node & int \\
			\bottomrule
		\end{tabular}
		\label{t1}
	\end{table}
	
	\subsection{Enhanced Cooperative Graph Neural Networks}\label{III-2}
    Prior QoR prediction methods such as HARP \cite{Sohrabizadeh} and HGBO \cite{Kuang} employ $conv$ layers with limited expressive capacity, where even complex graph representations or increased network depth yield limited prediction accuracy improvements. In contrast, CoGNN \cite{Finkelshtein} dynamically learns the topological structure of CDFGs according to training objectives, enabling more accurate predictions while maintaining stronger generalization capability. To enhance CoGNN's capability in capturing information flow between edges in multidirectional graphs, we developed ECoGNN as a specialized graph encoder for QoR prediction. We define the set of node states $\Phi$ as follows:
    \begin{equation}
    	\Phi = \{S, L_{in}, L_{out}, B, I\}
    \end{equation}
    where $S$, $B$, and $I$ denote the node states for information aggregation: $S$ receives neighbor information while broadcasting its own state, $B$ exclusively broadcasts its own state without receiving neighbor information, $I$ neither broadcasts nor receives information; $L_{in}$ receives exclusively from neighbors connected via incoming edges, and $L_{out}$ receives exclusively from neighbors connected via outgoing edges. 
    	
    MPNNs treat all nodes as possessing equivalent states during message passing, corresponding to the standard states $S$ defined in our action set. In contrast, ECoGNN dynamically creates edge weights (0 or 1) based on edge features to selectively control each node’s state evolution throughout the message-passing process. Fig. \ref{nc} demonstrates the distinct node behaviors between conventional MPNNs and ECoGNNs during message propagation. At $\zeta$=0, both models initialize each node's state to a standard state. At $\zeta$=1, MPNNs’ nodes retain the standard $S$ state throughout message passing, while ECoGNNs dynamically configure node states to enable feature updates tailored to task-specific characteristics. For example, at $\zeta$=1, the states of node $u$ and node $v$ are $B$ and $L_{in}$ respectively; at $\zeta$=2, their states change to $L_{in}$ and $B$. By endowing nodes with distinct states during message propagation, the inherent over-smoothing issue and limited expressive capacity of MPNNs can be effectively mitigated.
    
    A complete ECoGNN ($\pi$, $\eta$) layer consists of action networks $\eta$ and environment networks $\pi$. Action networks $\pi$ help each node determine the way information flows in and out, while environment networks $\eta$ update the features of each node according to the state of the node. Therefore, ECoGNNs update the node feature representation $h_c$ of each node $c$ according to the following steps. To begin with, the action networks $\pi$ will obtain the action of each node. For a set of actions $\Phi = \{S, L_{in}, L_{out}, B, I\}$, the action of node $c$ can be drawn from a categorical distribution $p_c\in\mathbb{R}^{|\Omega|}$ by aggregating the information of its neighbors $N(c)$. This process can be formulated as follows:
    
    \begin{equation}
    	p_c = \pi\left(h_c, N(c)\right) 
    	\label{5}
    \end{equation}
    
    Next, we input the probability vector $p_c$ into the Gumbel-softmax Estimator \cite{Jang2017} to obtain the node's action $a_c$. We can use environment networks $\eta$ to update the feature representation $h_c$ and obtain $h_{c}^{'}$. This process can be formulated as follows:
	
	\begin{equation}
		h_{c}^{'}= \eta\left(h_c, \mathcal{N}\right) 
		\label{6}
	\end{equation}
	
	Where $\mathcal{N}$ is related to the value of $a_c$:
	
	\begin{equation}
		\mathcal{N} = \begin{cases}
			\left\lbrace h_q | q \in N(c), a_q = S \lor B \right\rbrace, & a_c= L_{in} \lor L_{out} \lor S
			\\ \left\lbrace{}\right\rbrace, & a_c = B \lor I
		\end{cases}
		\label{7}
	\end{equation}
   	
   	\begin{figure*}[t]
   		\centering
   		\includegraphics[width=\linewidth]{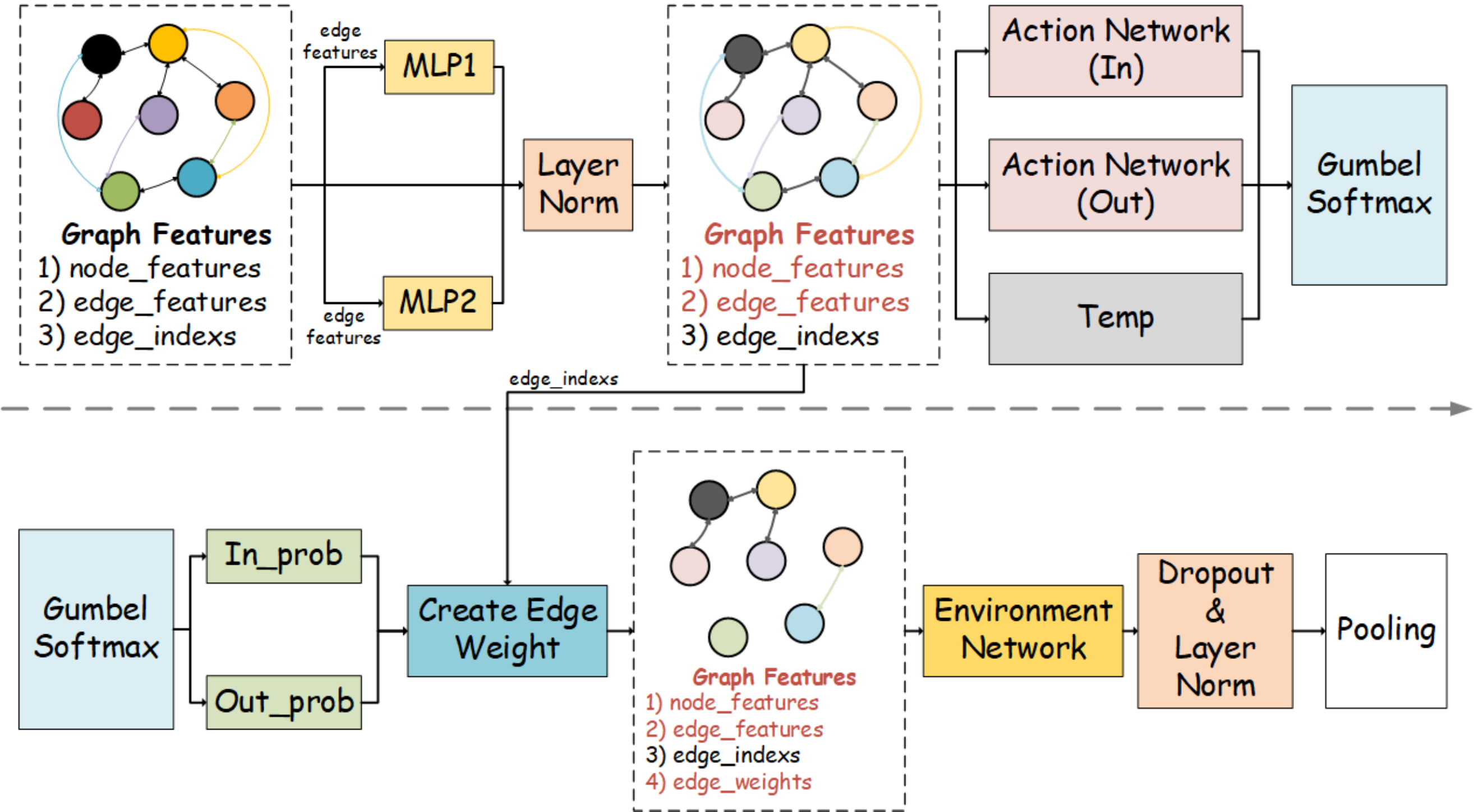}
   		\caption{Schematic diagram of ECoGNN's internal structure.}
   		\label{pIs}
   		\vspace{-3mm}
   	\end{figure*}
   	
   	We have provided an abstract representation of the message-passing process of ECoGNN, and the details of our predictive model will be further explained in the following content. We assume that both the Environment network and the Action network have only one layer, and the model structure diagram is shown in Fig. \ref{pIs}. To begin with, we input node features $h^{'}$ into LayerNorm \cite{Ba2016} to obtain the regularized feature $h$ and edge features are projected into two distinct dimensional spaces to be respectively fed into the Action network and Environment network. Since node actions follow a categorical distribution that renders the training process non-differentiable. We employ the Gumbel-Softmax estimator to achieve differentiable training. We consider a set of actions $\Phi$, which can be represented as a probability vector $p\in\mathbb{R}^{|\Phi|}$ and the vector $p$ stores the probability values of different actions. Let $w$ be a a categorical variable with class probabilities $p_1, p_2,..., p_{|\Phi|}$ and we assume categorical samples are encoded as k-dimensional one-hot vectors. The Gumbel-Max is a special reparametrization trick that provides an efficient way to draw samples $b$ from a categorical distribution with class probabilities $p$:
    \begin{equation}
   		b = onehot\left(\underset{k}{argmax}\left( g_k + \log{p_k}\right)\right)
   	\label{2}
    \end{equation}
   
   where $g_1, g_2,..., g_{|\Phi|}$ are samples drawn from Gumbel$(0,1)$. We can use the softmax function to replace the non-differentiable argmax function to generate the Gumbel-softmax scores $s_k$ and $\tau$ is the softmax temperature:
   
   \begin{equation}
   	s_k = \cfrac{\exp\left( \left(g_k + \log{p_k}\right)/\tau\right) }{\sum^{|\Phi|}_{q=1}\exp\left( \left(g_q + \log{p_q}\right)/\tau\right)}
   	\label{3}
   \end{equation}
   
	The Environment network and Action network are implemented using some MPNNs, with Temp (learnable Gumbel softmax temperature) primarily realized via MLPs. The weights of the edges $edge\_weights$ are constructed for outgoing and incoming edges based on the index of the edges $edge\_indexs$  and the probability of information inflow and outflow $prob_{in}$, $prob_{out}$, and are integrated with the $\mathcal{G}_{f}^{'}$ and fed into the Environment network to yield $h_{f}^{'}$, which is then regularized via LayerNorm. We obtain the final node-level representations $h_{f}$. To derive graph-level embeddings $h_{\mathcal{G}}$, these node-level representations are directly fed into an MLP for QoR prediction. However, using average pooling causes significant information loss. To address this, we leverage global node attention \cite{Li1} to generate the final graph-level embedding:
	\begin{equation}
		h_{\mathcal{G}} = \sum_{z=1}^{n} \left( \cfrac{\exp^{MLP\left(h_{f}^z\right) }}{\sum_{m=1}^{n}\exp^{MLP\left(h_{f}^m \right) }}\right) \times MLP\left(h_{f}^z\right)
		\label{8}
	\end{equation}
	where $n$ is the total number of nodes in the data graph, and $MLP$ is used to map $h_{f}^z$ and $h_{f}^m$ to $\mathbb{R}$. 
	
	Finally, the graph-level embeddings $h_{\mathcal{G}}$ are fed into an MLP prediction head to yield predicted results. The root mean squared error (RMSE) loss is computed to update model parameters. While mean absolute percentage error (MAPE) also reflects prediction accuracy, for targets like latency with values exceeding hundreds of thousands, a 1\% error translates to thousands of clock cycles. Thus, RMSE more accurately quantifies prediction error.

	\subsection{Large Language Model aided Meta-Heuristic Algorithms}\label{III-3}
    
	As discussed in Section \ref{icl}, pre-trained LLMs possess strong in-context learning capabilities. The DSE task is essentially a constrained multi-objective optimization problem. By incorporating relevant domain knowledge and exemplary input-output pairs into the in-context prompt, LLMs can leverage their acquired training knowledge to assist metaheuristic algorithms in navigating the design space. 

	\begin{figure*}[t]
		\centering
		\includegraphics[width=\linewidth]{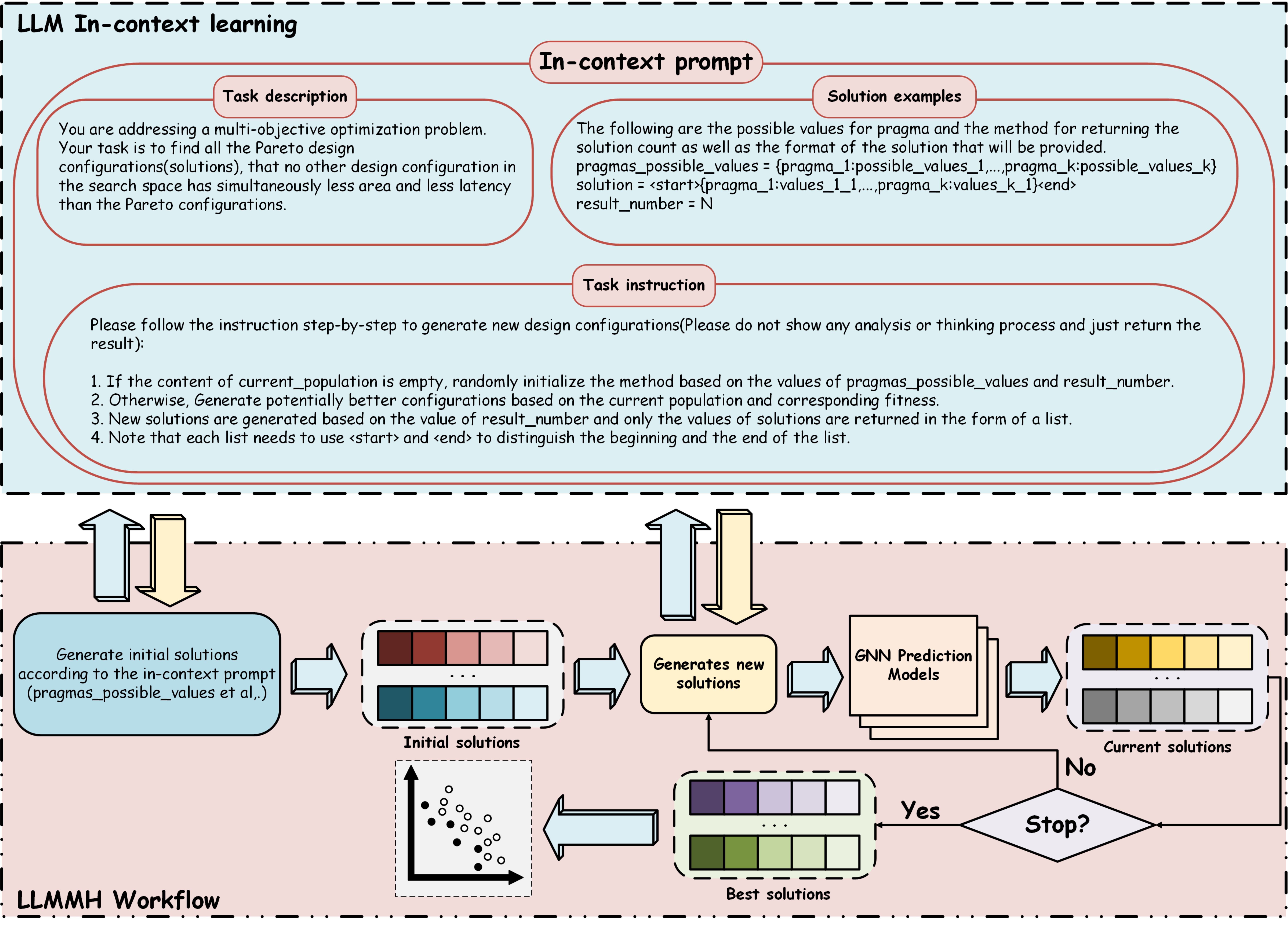}
		\caption{The LLMMH framework fully leverages the in-context learning capability of LLMs. By constructing an in-context prompt, it guides LLMs to understand the DSE task and further assists metaheuristic algorithms in generating new solutions. Specifically, the in-context prompt comprises task descriptions, solution examples, and task instructions. This workflow consists of initialization, LLM-guided solution generation, and GNN-based evaluation. In this study, we present and evaluate three LLM-driven meta-heuristics: LLMGA, LLMSA, and LLMACO.}
		\label{LLMMH}
		\vspace{-3mm}
	\end{figure*}
    
	As illustrated in Fig. \ref{LLMMH}, the LLMMH framework is structured into two primary components. First, the LLMMH utilizes LLMs' in-context learning capability through the OpenAI API \footnote{https://openai.com/}, where designed in-context prompts guide the LLMs to execute specific DSE steps. The in-context prompt consists of three components: task description, solution examples, and task instruction. Among these, task description is used to provide LLMs with knowledge related to DSE. Solution examples are used to present input-output instances (the input is usually the names and possible values of pragma directives, while the output is a set of pragma directive values). And task instruction is used to prompt LLMs on the process of generating solutions, so as to reduce the likelihood of LLMs outputting incorrect results. Second, within the workflow of LLMMH, it initiates the process by collecting pragma directive information to construct an in-context prompt that guides the LLMs in generating an initial solution. During the first iteration, the framework utilizes this pre-built prompt to produce new solutions, which are evaluated alongside the initial solution via the GNN predictive model. It then determines the current optimal solution and assesses whether the termination condition (maximum number of explorations) are met. If satisfied, the optimal solution is outputted. Otherwise, the procedure iteratively returns to the solution generation phase until convergence. Compared with the random operators of traditional metaheuristic algorithms, those of LLMMH are more characterized by "heuristics".
	
	{\bf{Why not employ fine-tuning?}} First, our goal is to improve the quality of exploration results of metaheuristic algorithms, rather than having LLMs handle DSE entirely. Metaheuristic algorithms require manual parameter setting based on benchmarks and are highly dependent on the parameters shown in Table \ref{ap}. These parameters are mainly used to generate new solutions. To address this issue, we consider using LLMs to generate new solutions. Thus, in the proposed LLMMH framework, the three LLM-guided metaheuristic algorithms do not require manual setting of these parameters at all. In fact, as long as sufficient knowledge and example pairs are provided in the in-context prompt, it is feasible to solve DSE problems using only the in-context learning capability of LLMs without relying on fine-tuning. Second, fine-tuning LLMs for DSE incurs high computational costs. Under the constraint of limited computational costs, even if only language models with several million parameters are fine-tuned, their performance may not necessarily be better than that of LLMs with a certain level of in-context learning capability.

	\begin{algorithm}[t]\footnotesize
		\caption{LLMGA algorithm.}
		\label{LLMEA}
		\renewcommand{\algorithmicrequire}{\textbf{Input:}}
		\renewcommand{\algorithmicensure}{\textbf{Output:}}
		\begin{algorithmic}[1]
			\REQUIRE Design space $ds$, design space size $S$, pragma possible values $V_p$, maximum number of explorations $N_{se}$, population size $N$, graph code $code$, optimal solution stagnation threshold $N_{ost}$, LLM temperature decay rate $v_{d}$, LLM temperature $t$; 
			\ENSURE Pareto optimal solution set $P_{est}$;
			\STATE $n = 1, n_{ost} = 0$; 
                \STATE $prompts \leftarrow$ employing prompt engineering techniques based on $N$, $V_p$ to establish LLM prompt;
			\STATE $P \leftarrow$ use $prompts$ to instruct LLM to initialize $N$ solutions; 
                \STATE 	$Fitness \leftarrow$ Evaluator $(P, code)$;
			\WHILE{$n \leq N_{se}$}
			\STATE $prompts \leftarrow$ employing prompt engineering techniques based on $N$, $V_p$, $Fitness$, $P$ to establish LLM prompt;
			\STATE $P_{new} \leftarrow$ use $prompts$ to instruct LLM to generate new offspring solutions;	
			\STATE $P, Fitness, n_{ost}\leftarrow$ Selector$(P, P_{new}, code, n_{ost})$; 
			
			\IF{$n_{ost} \geq N_{ost}$, $t \geq 0$}
			\STATE $t \leftarrow t - v_{d}$ ;
			\ENDIF
			\STATE $n \leftarrow n + N$	;
			\ENDWHILE
			\STATE $P_{est} \leftarrow$ the best solutions in $P$;
			\RETURN $P_{est}$
		\end{algorithmic}
	\end{algorithm}

	Algorithm \ref{LLMEA} provides an overview of the proposed LLM-enhanced genetic algorithm (LLMGA). The inputs of the algorithm include the parameters required for LLM prompts and those related to the design space, while the output is the Pareto-optimal set $P_{est}$. First we construct $prompts$ based on parameters $N$ and $V$ (lines 2-3), then instruct LLMs to generate an initial population, which is evaluated by a pre-trained model to obtain $Fitness$ (line 4). The process then enters an iterative phase: prompts are dynamically rebuilt using current $P$ and $Fitness$ to guide LLMs in generating new solutions $P_{new}$. The exploration engine employs the pre-trained predictive models to assess the $Fitness$ of each new solution in the current population and selects top configurations, updating both the current population $P$ and returning the fitness of population $Fitness$ (line 8). The fitness is primarily composed of two key metrics: estimated latency derived from critical path analysis and resource utilization metrics including LUT, DSP, FF and BRAM usage. The temperature parameter in LLMs governs the stochasticity of text generation during sampling: elevated temperature values induce higher randomness in output generation, while reduced temperature settings promote deterministic responses. This mechanism directly mirrors the exploration-exploitation trade-off inherent in DSE optimization processes. To better balance the exploration-exploitation trade-off, we implement an adaptive temperature mechanism specifically through dynamic updating of LLMs' temperature parameters (lines 9-11) \cite{Brown}. This process repeats until the loop terminates, at which point the solutions in $P$ are added to $P_{est}$, and the Pareto-optimal solution set $P_{est}$ is returned (lines 14-15). 

	\begin{algorithm}[t]\footnotesize
		\caption{LLMSA algorithm.}
		\label{LLMSA}
		\renewcommand{\algorithmicrequire}{\textbf{Input:}}
		\renewcommand{\algorithmicensure}{\textbf{Output:}}
		\begin{algorithmic}[1]
			\REQUIRE Design space $ds$, design space size $S$, pragma possible values $V_p$, graph code $code$, optimal solution stagnation threshold $N_{ost}$, LLM temperature decay rate $v_{d}$, LLM temperature $t$, population size $N$, initial temperature $T_{i}$, stop iteration temperature $T_{s}$, cooling rate $r_{cs}$; 
			\ENSURE Pareto optimal solution set $P_{est}$;
			\STATE $n_{t} = T_{i}, n_{ost} = 0$;
                \STATE $P, Fitness \leftarrow$ establish $prompt$ to instruct LLM to initialize $N$ solutions and evaluate each solution in $P$;
			\WHILE{$n_{t} >= T_{s}$}
			\STATE $P_{temp}, fitness, n_{ost} \leftarrow$ establish LLM $prompt$ to instruct LLM to generate neighbor solutions $P_{new}$, evaluate each solution in $P_{new}$ and select the top solutions into $P_{temp}$;
			\STATE $\Delta_{Cost} = Fitness_{avg} - fitness_{avg}$;
			\IF{$\Delta_{Cost}$}
			\STATE $P \leftarrow P_{temp}$, $Fitness \leftarrow fitness$ with $Prob = 1 - \exp^{\frac{\Delta_{Cost}}{n_{t}}}$;
			\ELSE 
			\STATE $P \leftarrow P_{temp}$, $Fitness \leftarrow fitness$;
			\ENDIF
			\IF{$n_{ost} \geq N_{ost}$, $t \geq 0$}
			\STATE $t \leftarrow t - v_{d}$ ;
                \ENDIF
                \STATE $n_{t} = \frac{n_{t}}{1 + r_{cs}}$;
			\ENDWHILE
			\STATE $P_{est} \leftarrow$ the best solutions in $P$;
			\RETURN $P_{est}$
		\end{algorithmic}
	\end{algorithm}
    
    Algorithm \ref{LLMSA} presents an overview of the LLM-enhanced simulated annealing algorithm (LLMSA). LLMSA maintains fundamental alignment with LLMGA in integrating LLMs in lines 2 and 4 (the only difference is that we instruct LLM to generate neighbor solutions as new solutions). The algorithm implements a conditional check on value $\Delta_{Cost}$: if $\Delta_{Cost} \textgreater 0$, it computes the probability for potentially accepting inferior solutions. If $\Delta_{Cost} \textless 0$, immediate acceptance occurs (indicating newly generated solutions surpass the current optimal set). This dual-branch mechanism enables dynamic balance between exploration ($\Delta_{Cost} \textgreater 0$ cases) and exploitation ($\Delta_{Cost} \textless 0$ scenarios) during optimization iterations (lines 5-10). The probability formula in line 11 exhibits temperature-dependent behavior: acceptance probability for inferior solutions increases with higher temperatures and decreases with lower thermal states.

     Algorithm \ref{LLMACO} presents an overview of the LLM-enhanced ant colony optimization algorithm (LLMACO). To preserve ACO's core characteristics, it retains the pheromone matrix mechanism. It initializes a pheromone matrix $P_{m}$ encoding concentration values for each pragma option (line 2). While maintaining similar LLMGA operations, the prompt construction phase explicitly injects pheromone matrix data into LLMs (lines 3-5). Before each iteration concludes, pheromone evaporation and concentration updates are applied to values in current elite solutions (lines 6-7). Subsequent procedures mirror LLMGA, crucially replacing traditional ACO's probabilistic selection formulas (requiring complex computations) with LLM-driven decision-making.
	
    \begin{algorithm}[t]\footnotesize
		\caption{LLMACO algorithm.}
		\label{LLMACO}
		\renewcommand{\algorithmicrequire}{\textbf{Input:}}
		\renewcommand{\algorithmicensure}{\textbf{Output:}}
		\begin{algorithmic}[1]
			\REQUIRE Design space $ds$, design space size $S$, pragma possible values $V_p$, maximum number of explorations $N_{se}$, the number of ants $N_{ant}$, graph code $code$, optimal solution stagnation threshold $N_{ost}$, LLM temperature decay rate $v_{d}$, LLM temperature $t$, pheromone evaporation ratio $\rho$; 
			\ENSURE Pareto optimal solution set $P_{est}$;
			\STATE $n = 1, n_{ost} = 0$; 
			\STATE Initializes a pheromone matrix $P_{m}$ based on the shape of $V_{p}$;
                \STATE $P, Fitness \leftarrow$ establish $prompt$ to instruct LLM to initialize $N$ solutions and evaluate each solution in $P$;
			\WHILE{$n \leq N_{se}$}
			\STATE $P, Fitness, n_{ost} \leftarrow$ establish LLM $prompt$ to instruct LLM to generate new solutions $P_{new}$ based on pheromone matrix $P_{m}$, evaluate each solution in $P_{new}$ and select the top solutions into $P$;
			\STATE Pheromone evaporation $P_{m} = \rho \times P_{m}$;
			\STATE Update pheromone matrix $P_{m}$  $\tau_{i}(c) = \tau_{i}(c) + 0.1$; 
			\IF{$n_{ost} \geq N_{ost}$, $t \geq 0$}
			\STATE $t \leftarrow t - v_{d}$ ;
			\ENDIF
			\STATE $n \leftarrow n + N_{ant}$	;
			\ENDWHILE
			\STATE $P_{est} \leftarrow$ the best solutions in $P$;
			\RETURN $P_{est}$
		\end{algorithmic}
	\end{algorithm}
    
	\section{Experiments}
	\subsection{Experimental Setup}
	In the experiments, the datasets are sourced from SOTA works GNN-DSE \cite{Sohrabizadeh} and HGBO-DSE \cite{Kuang}. The GNN-DSE dataset, containing latency and resource usage labels, is extracted from HLS reports, while the HGBO-DSE dataset with PPA results is derived from post-implementation reports. Since the proposed method can be used for predicting both the post-HLS QoRs and the post-implementation QoRs, we compare it with SOTA works on the aforementioned two datasets respectively. To evaluate the proposed DSE approach, we use the most accurate prediction model based on experimental evaluation as the estimator for DSE, which collaborates with LLMMH in Section \ref{III-3} to ultimately obtain the Pareto-optimal solution set. The applications in the GNN-DSE dataset are sourced from MachSuite \cite{Reagen} and Polyhedral \cite{Yuki} benchmarks, which involve 7 applications for training, testing, and validation and 5 for DSE exploration, while the HGBO-DSE dataset includes 10 applications from MachSuite, with 6 for training, testing, and validation and 4 for inference. The experiments are conducted on the AMD Ultrascale+ MPSoC ZCU104. The benchmarks are synthesized using Vitis-HLS 2022.1 and Vivado 2022.1 to collect ground-truth latency and resource utilization metrics, which serve as training labels for the model. 
	
	Both datasets are randomly divided into 70\% for training, 15\% for testing, and 15\% for validation. In the experiments, ECoGNNs' environment network has three layers and the action network has two layers with MEANGNNs, SUMGNNs \cite{Hamilton}, and GCN \cite{Kipfetal} as the main GNN types selectable GNN types for both networks, from which the most suitable combination for the QoR prediction task is chosen. Meanwhile, the hidden layer dimension of all models is set to 128, and the Adam optimizer is used, with the GNN-DSE dataset experiments using a batch size of 64, a learning rate of 0.001, and over 500 iterations, while the HGBO-DSE dataset experiments employ a batch size of 128, a learning rate of 0.001, and over 250 iterations. To be consistent with GNN-DSE \cite{Sohrabizadeh} and HGBO-DSE \cite{Kuang}, Root Mean Squared Error (RMSE) loss is used to quantify the post-HLS QoR prediction quality, and Mean Absolute Error (MAE) and Mean Absolute Percentage Error (MAPE) are used to evaluate the accuracy of the models for the post-implementation QoR prediction.

	\subsection{Evaluation of Post-HLS QoR Prediction Accuracy}\label{eval 2}
	
	In this experiment, to validate the prediction accuracy of ECoGNNs for estimating post-HLS QoR metrics, we leverage the architectural flexibility of ECoGNNs to design eight combinations of environment and action networks for predicting latency, LUT, DSP, FF and BRAM usage, comparing their performance against baseline models. For clarity, we denote MEANGNN, SUMGNN, and GCN as $\alpha$, $\beta$, $\gamma$, respectively, to describe different ECoGNN configurations. The SUMGNNs and MEANGNNs mentioned here are two fundamental variants of MPNNs that differ in their neighborhood aggregation functions, with SUMGNNs using summation-based aggregation and MEANGNNs employing mean-based aggregation. Additionally, we evaluate baseline models including GNN-DSE \cite{Sohrabizadeh},  HGP+SAGE+GF \cite{Kuang}, IronMan-PRO \cite{Wu2}, PNA-R \cite{Wu1}, and PowerGear \cite{Lin}. While PNA-R and PowerGear are primarily designed for post-implementation prediction tasks (PowerGear proposes GNN to predict power), we adapt these baseline models for post-HLS scenarios by modifying their input dimensions to align with the feature space of the GNN-DSE dataset and restructuring their architectures using the PyTorch Geometric framework. As shown in Table \ref{table 1}, the HLS dataset utilizes 9 benchmarks, comprising a total of 4320 CDFGs. Among these, CDFGs from 6 benchmarks including \textit{aes}, \textit{gemm-blocked}, \textit{gemm-ncubed}, \textit{spmv-crs}, \textit{spmv-ellpack} and \textit{nw} are allocated for training, while the remaining 3 benchmarks' CDFGs are reserved for validation and inference.

    \begin{table*}
		\renewcommand{\arraystretch}{1.2}
		\setlength{\tabcolsep}{3pt}
		\tiny
		\centering
		\caption{The Training, Validation and Inference Applications Used for Post-HLS QoR Prediction.}
		\begin{tabular}{l|c|c|c}
			\toprule
			\bf{Kernel name} & \bf{Description} & \bf{\# Pragmas} & \bf{\# Design configs} \\
			\midrule
			\textit{aes} & a common block cipher & 3 & 45 \\
			\hline
            \textit{atax}&Matrix Transpose and Vector Multiplication&5&3,354\\
            \hline
			\textit{gemm-blocked} & A blocked version of matrix multiplication & 9 & 2,314\\                              
            \hline
            \textit{gemm-ncubed} & O(n3) algorithm for dense matrix multiplication. & 7 & 7,792\\
			\hline
			\textit{mvt}&Matrix Vector Product and Transpose&8&3,059,001\\                     
			\hline
			\textit{spmv-crs} & Sparse matrix-vector multiplication, using variable-length neighbor lists& 3 & 114 \\
			\hline
			\textit{spmv-ellpack} & Sparse matrix-vector multiplication, using fixed-size neighbor lists & 3 & 114 \\
			\hline
			\textit{stencil3d} & A three-dimensional stencil computation & 7 & 7,591 \\
            \hline
            \textit{nw} & A dynamic programming algorithm for optimal sequence alignment& 6 & 15,288 \\
			\bottomrule
		\end{tabular}
		\label{table 1}
		
	\end{table*}
	
	\begin{table}
		\renewcommand{\arraystretch}{1.1}
		\setlength{\tabcolsep}{3pt}
		\centering
		\caption{RMSE Loss of Different ECoGNNs on Unseen Applications. The Top Models are Marked in Bold.}
		\label{table 2}
		\begin{tabular}{l|c|c|c|c|c|c}
			\toprule
			{\bf{Model}}&{\bf{Latency}}&{\bf{LUT}}&{\bf{DSP}}&{\bf{FF}}&{\bf{BRAM}}&{\bf{All}} 
			\\
			\midrule
			{ECoGNNs($\alpha$,$\alpha$)}&0.4315&0.0043&0.0114&0.0155&0.0270&0.4897
			\\
			\hline
			{ECoGNNs($\alpha$,$\beta$)}&0.7101&0.0084&0.0141&0.0279&0.0333&0.7937 
			\\
			\hline
			{ECoGNNs($\beta$,$\beta$)}&0.5077&0.0053&0.0091&\bf{0.0146}&0.0353&0.5720 
			\\
			\hline
			{ECoGNNs($\beta$,$\alpha$)}&{\bf{0.3557}}&\bf{0.0039}&\bf{0.0075}&0.0152&\bf{0.0244}&\bf{0.4067}
			\\
			\hline
			{ECoGNNs($\gamma$,$\beta$)}&0.6748&0.0067&0.0167&0.0229&0.0331&0.7541
			\\
			\hline
			{ECoGNNs($\gamma$,$\gamma$)}&0.5825&0.0063&0.0126&0.0252&0.0324&0.6592
			\\
			\hline
			{ECoGNNs($\alpha$,$\gamma$)}&0.4131&0.0174&0.0276&0.1193&0.0297&0.6072
			\\
			\hline
			{ECoGNNs($\beta$,$\gamma$)}&0.6456&0.0112&0.0248&0.0434&0.0346&0.7596
			\\
			\bottomrule
		\end{tabular}
         \vspace{-3mm}
	\end{table}	

	As shown in Table \ref{table 2}, we conducted extensive experiments on this dataset and recorded the RMSE achieved by each model. RMSE is squared-error metric measuring prediction accuracy via root-averaged discrepancies between predicted and actual values, preserving data units. For latency prediction, ECoGNNs($\beta$, $\alpha$) achieves the best performance, followed by ECoGNNs($\alpha$, $\gamma$) with prediction errors of 0.3557 and 0.4131, respectively. For DSP and FF resource prediction, ECoGNNs($\beta$, $\alpha$) and ECoGNNs($\beta$, $\beta$) achieve the better results. For the prediction of LUT and BRAM, ECoGNNs($\beta$, $\alpha$) and ECoGNNs($\beta$, $\beta$) achieve the lowest prediction errors, respectively. Although ECoGNNs($\beta$, $\alpha$) has a slightly higher prediction error (0.0006) than ECoGNNs($\beta$, $\beta$) in FF prediction, its overall error is 0.083 lower than ECoGNNs($\alpha$, $\alpha$). Thus, we conclude that ECoGNNs($\beta$, $\alpha$) exhibits the best ECoGNN combination on the GNN-DSE dataset.

	\begin{table}
		\renewcommand{\arraystretch}{1.1}
		\setlength{\tabcolsep}{3pt}
		\centering
		\caption{RMSE Loss of ECoGNNs($\beta$,$\alpha$) and SOTA Models on Unseen Applications.}
		\label{eval 3}
		\begin{tabular}{l|c|c|c|c|c|c}
			\toprule
			{\bf{Model}}&{\bf{Latency}}&{\bf{LUT}}&{\bf{DSP}}&{\bf{FF}}&{\bf{BRAM}}&{\bf{All}} 
			\\
			\midrule
			{GNN-DSE\cite{Sohrabizadeh}}&0.5359&0.0762&0.1253&0.0632&0.0515&0.8521\\
			\hline
			{HGP+SAGE+GF\cite{Kuang}}&0.9519&0.0152&0.0270&0.0895&0.0362&1.1197\\
			\hline
			{IRONMAN-PRO\cite{Wu2}}&0.6778&0.0081&0.0161&0.0399&0.0326&0.7745\\
			\hline
			{PNA-R\cite{Wu1}}&1.3728&0.0144&0.0307&0.0811&0.0610&1.5600\\
			\hline
			{PowerGear\cite{Lin}}&1.3956&0.0177&0.0290&0.0908&0.0432&1.5764\\
			\hline
			{ECoGNNs($\beta$,$\alpha$)}&{\bf{0.3557}}&\bf{0.0039}&\bf{0.0075}&\bf{0.0152}&\bf{0.0244}&\bf{0.4067}
			\\
			\bottomrule
		\end{tabular}
            \vspace{-3mm}
	\end{table}
	
	In prior experiments, the combination of ECoGNNs components significantly influences prediction accuracy. We observe that when the environment and action networks utilize simpler MPNNs like SUMGNN and MEANGNN, prediction errors are substantially lower compared to using more complex models such as GCN. This phenomenon is fundamentally tied to ECoGNNs' operational mechanism (it dynamically selects graph topologies to regulate message flow between nodes). In contrast, complex MPNNs like GCN inherently prioritize the importance of information from neighboring nodes, which contradicts ECoGNNs' design philosophy. ECoGNNs autonomously learn which neighbors' information is relevant for each node, effectively disregarding messages from unimportant neighbors. Consequently, integrating MPNNs like GCN, which predefine neighbor importance, results in suboptimal performance within the ECoGNNs framework.

\begin{figure*}[t]
		\centering
		\includegraphics[width=0.9\linewidth]{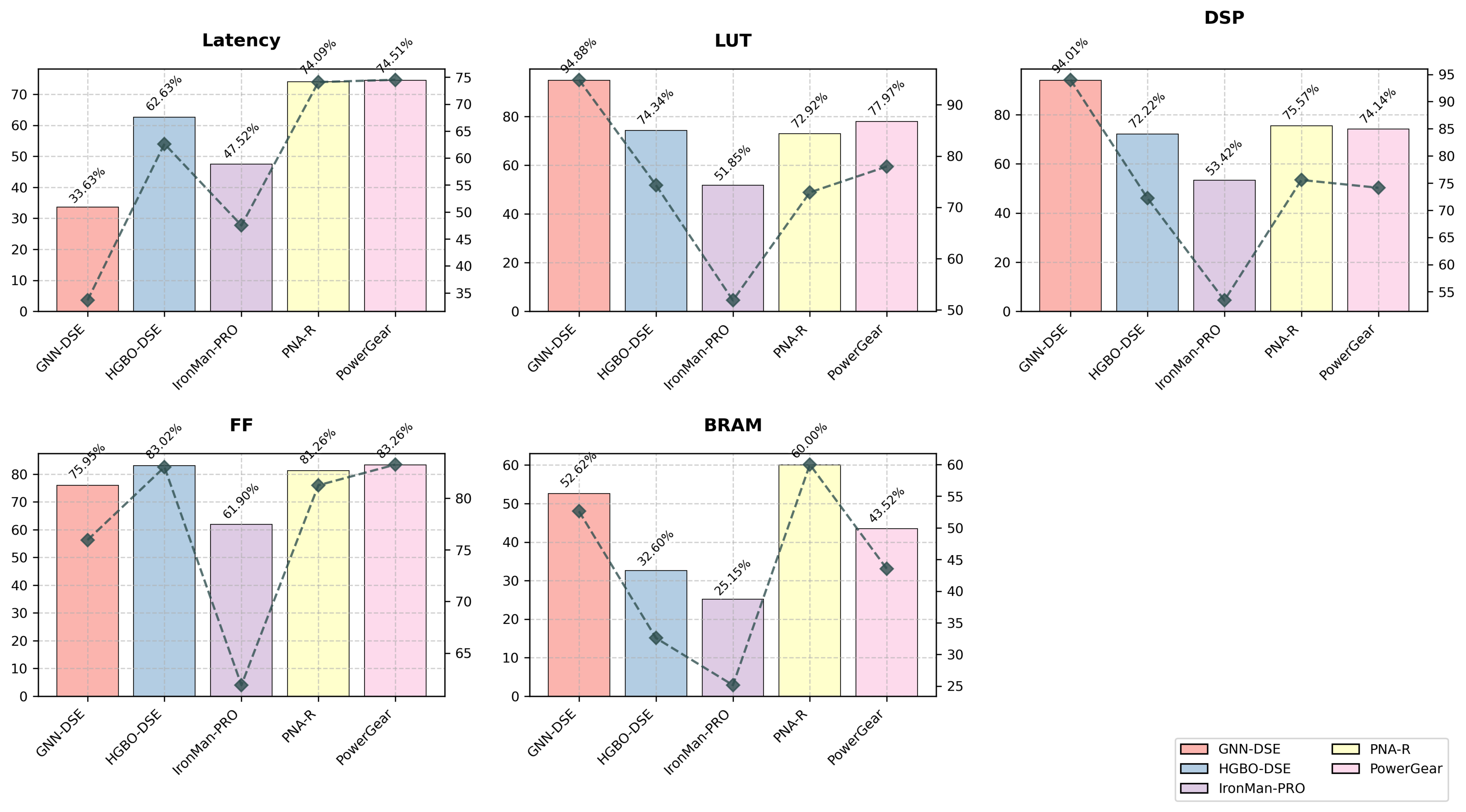}
		\caption{Reduction rates of RMSE loss achieved by ECoGNNs($\beta$,$\alpha$) over baseline models. The bar chart quantifies each model’s prediction error reduction ratios (PERR) relative to the top-performing model, while the line chart tracks the evolving PERR trend across evaluation metrics.}
		\label{comp baseline}
         \vspace{-5mm}
	\end{figure*}

    As shown in Table \ref{eval 3}, ECoGNNs($\beta$, $\alpha$) also exhibits exceptional prediction capabilities compared to SOTA models. Compared to GNN-DSE,  HGP+SAGE+GF, IRONMAN-PRO, PNA-R, and PowerGear,  ECoGNNs($\beta$, $\alpha$) can reduce RMSE values by 47.69\%, 36.32\%, 52.51\%, 26.07\%, and 25.80\% respectively. This superior performance stems from ECoGNNs' task-specific feature adaptation. Unlike traditional MPNNs, which focus primarily on minimizing loss during training, ECoGNNs prioritize the extraction of graph-level and node-level features to achieve task objectives, with loss reduction being a natural consequence of this process. As evidenced by Fig. \ref{comp baseline}, ECoGNNs architectures demonstrate statistically significant improvements over baseline models. The ECoGNNs($\beta$, $\alpha$) variant exhibits superior performance across all evaluation metrics, particularly achieving at least 33.63\% RMSE reduction in latency measurements. With particular strength in hardware utilization metrics, the framework achieves peak error reduction of 94.01\% for DSP and mean error reductions of 74.44\% (LUT), 77.08\% (FF), and 42.78\% (BRAM) respectively, outperforming existing approaches in FPGA resource estimation accuracy.
	
	To further verify the performance of ECoGNNs, we also compare it with the SOTA work HARP \cite{Sohrabizadeh1}. While HARP proposes a hierarchical graph representation for HLS designs using auxiliary nodes, ECoGNN leverages a different graph representation aligned with GNN-DSE. To facilitate a direct comparison between these approaches, we have reconstructed the HARP dataset for ECoGNN by adopting the graph representation from GNN-DSE and integrating it with the original performance labels. These labels are derived from HARP's v1 (SDAccel 2018.3) and v2 (Vitis 2020.2) datasets. The experimental results are shown in Table \ref{ch}, ECoGNNs demonstrates significantly superior QoR prediction capability compared to HARP, achieving a 57\% total RMSE reduction on the v1 and v2 databases. As discussed in Section \ref{III-2}, while HARP augments CDFG with enriched information, its graph encoder remains constrained by the MPNN paradigm's inherent limitations, resulting in inadequate feature extraction from enhanced CDFGs. In contrast, ECoGNN dynamically adapts to graph topology based on task characteristics, employs context-aware message passing, and comprehensively captures discriminative features within CDFGs.
	
	\begin{table}
		\centering
		\caption{ECoGNN and HARP were Evaluated on the HARP Dataset, Maintaining Consistent Prediction Targets with Previous Experiments and Employing Total RMSE as the Prediction Error Metric.}
		\label{ch}
		\begin{tabular}{l|c|c}
			\toprule
			{\bf{Model}}&{\bf{v1 database}}&{\bf{v2 database}}\\
			\hline
			HARP(Parallel\&mergeasNPT
			+postGNNlayer)\cite{Sohrabizadeh1} & 0.974 & 0.679\\
			\hline
			ECoGNN($\beta$, $\alpha$) & 0.415(-57.39\%) & 0.291(-57.14\%)\\
			\hline
		\end{tabular}
		\vspace{-3mm}
	\end{table}

	\subsection{Evaluation of Post-Implementation QoR Prediction Accuracy}\label{eval 1}
      \begin{table}
		\renewcommand{\arraystretch}{1.2}
		\setlength{\tabcolsep}{3pt}
		\centering
            \tiny
		\caption{The Benchmarks Used for Post-Implementation QoR Prediction. }
		\begin{tabular}{l|c|c|c|c|c}
			\toprule
			\bf{Kernel name} & \bf{Description} & \bf{\# Func} & \bf{\# Array} & \bf{\# Loop} & \bf{\# Params} \\
			\midrule
			\textit{aes} & a common block cipher & 12 & 2 & 11 & 30 \\
			\hline
			\textit{bfs-bulk} & Data-oriented version of breadth-first search & 1 & 4 & 3 & 25\\
			\hline
			\textit{fft-strided} &Recursive formulation of the Fast Fourier Transform& 1 & 4 & 2 & 27\\
			\hline
			\textit{gemm-ncubed} & O(n3) algorithm for dense matrix multiplication. & 1 & 3 & 3 & 27\\
			\hline
			\textit{md-knn} &n-body molecular dynamics, using k-nearest neighbors to compute only local forces& 1 & 7 & 2 & 42\\
                \hline
			\textit{nw} & A dynamic programming algorithm for optimal sequence alignment& 1 & 6 & 7 & 54 \\
			\hline
			\textit{sort-radix} &Sorts an integer array by comparing 4-bits blocks at a time& 7 & 4 & 11 & 72\\
			\hline
			\textit{spmv-ellpack} & Sparse matrix-vector multiplication, using fixed-size neighbor lists & 1 & 4 & 2 & 25\\
			\hline
			\textit{stencil3d} & A three-dimensional stencil computation & 1 & 2 & 9 & 43 \\
			\hline
			\textit{viterbi} &A dynamic programing method for computing probabilities on a Hidden Markov model& 1 & 6 & 7 & 52 \\
			\bottomrule
		\end{tabular}
		\label{dataset 1}
         \vspace{-3mm}
	\end{table}
	To further validate the accuracy of the proposed ECoGNNs, we also evaluate its performance on post-implementation QoR prediction. For LUT, FF, CP and POWER, we employ MAPE as the loss function. For DSP and BRAM where zero-value occurrences exist, we adopt MAE to avoid division-by-zero errors during model optimization.The HGBO-DSE dataset provides labels for LUT, DSP, FF, BRAM, CP and POWER, all extracted from implementation reports. In this experiment, we compare the MAPE and MAE of ECoGNNs($\beta$, $\alpha$) on the HGBO-DSE dataset against baseline models including  HGP+SAGE+GF, IRONMAN-PRO, PowerGear and PNA-R. It is important to note that PowerGear was initially designed to predict power, while PNA-R was created to predict metrics such as LUT, FF, and CP. In accordance with the experiments conducted in \cite{Kuang}, we only present the results for the target metrics of PowerGear and PNA-R. Notably, the HGBO-DSE dataset has a node feature dimension of 15, significantly smaller than the GNN-DSE dataset's 153-dimensional features. To account for this, we reduce the number of iterations, while ensuring that each model undergoes at least 250 training epochs. It is worth noting that HGBO-DSE's work does not evaluate the model generalization, including for all the reproduced baseline models. In this experiment, we partition the dataset into two subsets comprising 10 applications. As shown in Table \ref{dataset 1}, six applications (\textit{aes, bfs, fft, gemm, md, nw}) are used for training, while the remaining applications are reserved as unseen applications for inference.

	\begin{table}
		\renewcommand{\arraystretch}{1.2}
		\setlength{\tabcolsep}{3pt}
		\centering
		\caption{MAPE and MAE of ECoGNNs($\beta$, $\alpha$) and SOTA Models on Unseen Applications. }
		\label{table 4}
		\begin{tabular}{l|c|c|c|c|c|c}
			\toprule
			&\multicolumn{4}{c|}{\bf{MAPE(\%)}} & \multicolumn{2}{c}{\bf{MAE}}\\
			\hline
			{\bf{Model}}&{\bf{LUT}}&{\bf{FF}}&{\bf{CP}}&{\bf{POWER}}&{\bf{DSP}}&{\bf{BRAM}}\\
			\hline
			HGP+SAGE+GF\cite{Kuang}&9.05&9.87&9.87&13.60&1.4282&0.1606\\
			\hline
			IRONMAN-PRO\cite{Wu2}&7.35&6.30&10.98&16.34&0.6198&0.1917\\
			\hline
			PowerGear\cite{Lin}&-&-&-&11.61&-&-\\
			\hline
			PNA-R\cite{Wu1}&\bf6.40&6.52&14.03&-&-&- \\
			\hline
			{ECoGNNs($\beta$, $\alpha$)}&6.82&\bf5.98&\bf7.54&\bf10.01&\bf0.5339&\bf0.1035\\
			\bottomrule
		\end{tabular}
        \vspace{-5mm}
	\end{table}
	
	As shown in Table \ref{table 4}, it can be observed that ECoGNNs($\beta$,$\alpha$) achieves the smallest prediction errors for FF, CP, POWER, DSP, and BRAM compared to baseline models, with values of 5.98\%, 7.54\%, and 10.01\% in terms of MAPE, 0.5339 and 0.1033 in terms of MAE, respectively. For LUT prediction, PNA-R exhibits the lowest error at 6.40, while ECoGNNs($\beta$,$\alpha$) closely follows with 6.82, demonstrating a marginal difference. Notably, in POWER prediction, ECoGNNs($\beta$,$\alpha$) reduces the error to 10.01\%, outperforming PowerGear, a model specifically designed for power prediction, which achieves 11.61\%. The performance of  HGP+SAGE+GF on unseen applications underscores that complex model architectures or mere structural stacking cannot enhance performance in HLS or implementation prediction tasks. In summary, the task-adaptive ECoGNNs demonstrate exceptional performance in predicting post-place-and-route resource usage and exhibits strong generalization capabilities.
	
	Compared to SOTA works, ECoGNNs demonstrates superior performance across diverse scenarios including post-HLS and post-implementation QoR prediction. Experimental results reveal that improving the expressive power of graph encoders and mitigating inherent limitations of MPNNs such as over-smoothing can significantly improve the QoR prediction precision.

	\subsection{Evaluation of Design Space Exploration}
 the DSE experiment, we applied our proposed LLMMH framework to explore five unseen applications (\textit{atax, doitgen, gemm-p, heat-3d, and mvt}) that were excluded from the training phase, utilizing the ECoGNNs($\beta$, $\alpha$) model trained in Section \ref{eval 1} as the evaluator. Table \ref{table 5} presents the detail information of the five applications. For each application, the design configurations encompass all possible combinations of the pragma values and the C/C++ code. We implemented a Python-based explorer to convert these configurations into graph data via LLVM and ProGraML, enabling input into the prediction model. The evaluation results are stored, and the top-ranked configurations are identified through iterative refinement. The primary objective of DSE is to identify the Pareto-optimal configuration set. To quantify the quality of the approximate Pareto-optimal set, we employ the ADRS metric \cite{Schafer}:	
	\begin{equation}
		ADRS(\Gamma,\Omega) = \cfrac{1}{|\Gamma|}\ \underset{\lambda \in \Gamma}{\sum}\ \underset{\mu \in \Omega}{\min}f\left(\lambda, \mu \right)
		\label{9}
	\end{equation}
	
	\begin{equation}
		f(\lambda, \mu) = \max\left(|\frac{A(\lambda) - A(\mu)}{A(\mu)}|, |\frac{L(\lambda) - L(\mu)}{L(\mu)}|\right)
            \label{10}
	\end{equation}

        \begin{equation}
            A(j) = \frac{1}{4} * \left( \frac{FF_{j}}{FF_{max}} + \frac{LUT_{j}}{LUT_{max}} + \frac{BRAM_{j}}{BRAM_{max}} + \frac{DSP_{j}}{DSP_{max}} \right)
            \label{11}
        \end{equation}

        \begin{equation}
            L(j) = \frac{1}{Latency_{j}}
            \label{12}
        \end{equation}
        
    where $\Gamma$ represents the reference Pareto-optimal set, $\Omega$ denotes the approximate Pareto-optimal set, and the function $f$ computes the distance between $\lambda$ and $\mu$. Equations \ref{11} and \ref{12} formally define the functionality of A(·) and L(·) respectively, where $FF_{max}$, $LUT_{max}$, $BRAM_{max}$, and $DSP_{max}$ represent the maximum available resources on the target FPGA platform. The framework quantifies design performance through latency measurements as the primary evaluation metric. A lower ADRS value indicates a higher accuracy of the approximate Pareto-optimal set relative to the reference. To assess the performance of LLMMH variants (LLMGA, LLMSA and LLMACO) , we compare them with three meta-heuristic algorithms: NSGA-II \cite{Schafer1}, SA \cite{Schafer2}, ACO \cite{Schafer3}, as referenced in \cite{Wang}. The reference Pareto-optimal set is collected from the results generated by the exhaustive DSE algorithm proposed in \cite{Sohrabizadeh}. 
    
	To ensure fair comparison of algorithm performance, the parameter configurations for SA, NSGA-II, ACO and LLMMH variants maintain unified settings: both population size $N$ and maximum number of explorations $N_{se}$ are set to identical values across all algorithms. Additionally, due to the excessively large design space sizes of certain benchmarks such as \textit{mvt}, we have imposed a one-hour exploration time limit for each benchmark. The adaptive parameter scheme is dynamically adjusted based on the design space size $S$ as follows:
	\begin{equation}
		(N_{se}, N) =
		\begin{cases}
			(0.00005S, 30) & S \geq 10^7 \\
                (0.005S, 30) & 10^5 \textless S \leq 10^6 \\
                (0.05S, 30) & 10^4 \textless S \leq 10^5 \\
			(0.3S, 30) & 500 \textless S \leq 10^4 \\ 
			(0.5S, 10) & S \leq 500
		\end{cases}
	\end{equation}

	\begin{table*}
		\renewcommand{\arraystretch}{1.2}
		\setlength{\tabcolsep}{7pt}
		\centering
		\small
		\caption{The Unseen Applications and the Number of Design Configurations Used in DSE.}
		\label{table 5}
		\begin{tabular}{l | c | c | c}
			\toprule
			{\bf{Benchmark}}&\bf{Description}&\bf{\# Pragmas}&\bf{\# Design configs}
			\\
			\midrule
			\textit{atax}&Matrix Transpose and Vector Multiplication&5&3,354
			\\
			\textit{doitgen}&Multi-resolution analysis kernel (MADNESS)&6&179
			\\
			\textit{gemm-p}&Matrix-multiply C=alpha.A.B+beta.C&8&409,905
             \\
			\textit{heat-3d}&Heat equation over 3D data domain&11&71,511
			\\
			\textit{mvt}&Matrix Vector Product and Transpose&8&3,059,001
			\\
			\bottomrule
		\end{tabular}
        \vspace{-3mm}
	\end{table*}

	\begin{table*}
		\renewcommand{\arraystretch}{1.2}
		\setlength{\tabcolsep}{3pt}
		\centering
		\footnotesize
		\caption{ADRS Results and Runtime on Unseen Applications.}
		\label{table 6}
		\begin{tabular}{l | c | c | c | c | c | c| c}
			\toprule
			\bf{Algorithm}&\bf{atax}&\bf{heat-3d}&\bf{doitgen}&\bf{gemm-p}&\bf{mvt}&\bf{Average ADRS}&\bf{Overall Runtime (s)}
			\\
			\midrule
			SA&0&0.4053&0.0669&0.5567&0.5620&0.3182&213
			\\
			\hline
			NSGA-II&0&0.3532&0&0.2936&0.5557&0.2405&132
                \\
			\hline
			ACO&0.0043&0.2946&0&0.6081&0.5138&0.2842&151
			\\
			\hline
			LLMGA(deepseek-r1)&0.0086&0.1483&0&0.1633&0.3844&0.1409&791
			\\
			\hline
			LLMGA(gpt-4o)&0&0.2085&0.0790&0.1837&0.4206&0.1784&431
			\\
			\hline
			LLMGA(gpt-4.1)&0.0029&0.3182&0&0.2258&0.1549&\bf0.1404&396
			\\
			\hline
			LLMGA(o3-mini)&0&0.2829&0.0717&0.2007&0.3223&0.1755&702
			\\
			\hline
			LLMSA(deepseek-r1)&0.0086&0.1053&
            0&0.1235&0.1668&\bf0.0808&820
			\\
			\hline
			LLMSA(gpt-4o)&0.0028&0.1481&0&0.1389&0.3619&0.1304&360
			\\
			\hline
			LLMSA(gpt-4.1)&0&0.3237&0&0.2519&0.2416&0.1635&411
			\\
			\hline
			LLMSA(o3-mini)&0.0028&0.2878&0.0245&0.2239&0.4565&0.1991&684
			\\
			\hline
			LLMACO(deepseek-r1)&0.0086&0&0&0.0096&0.1516&\bf0.0339&839
			\\
			\hline
			LLMACO(gpt-4o)&0&0&0&0.0361&0.4297&0.0932&393
			\\
			\hline
			LLMACO(gpt-4.1)&0&0.2096&0&0.2401&0.2131&0.1326&347
			\\
			\hline
			LLMACO(o3-mini)&0&0.1863&0&0.2135&0.4084&0.1616&638
			\\
			\hline
			\multicolumn{4}{c|}{\bf{LLMACO(deepseek-r1) Improv. over SA}}&\multicolumn{4}{c}{\bf{89.34\%}}
			\\
			\multicolumn{4}{c|}{\bf{LLMACO(deepseek-r1) Improv. over NSGA-II}}&\multicolumn{4}{c}{\bf{85.90\%}}
			\\
			\multicolumn{4}{c|}{\bf{LLMACO(deepseek-r1) Improv. over ACO}}&\multicolumn{4}{c}{\bf{88.07\%}}
			\\
			\bottomrule
		\end{tabular}
	\end{table*}
	
    In the experiments, we employ four foundation models (deepseek-r1 \cite{deepseek-R1}, gpt-4o \cite{gpt4o}, gpt-4.1 \cite{gpt4.1} and o3-mini \cite{o3-mini}) as surrogate models. In total, we evaluate 12 combinations of LLMMH (3 meta-heuristics $\times$ 4 LLMs). As shown in Table \ref{table 6}, traditional meta-heuristics (NSGA-II, SA, ACO) achieve ADRS exceeding 0.5 on the \textit{mvt} benchmark with large design space, aligning with our theoretical analysis of their exploration limitations. In contrast, our LLMMH framework demonstrates superior performance—LLMACO (deepseek-r1) achieves ADRS as low as 0.0339. On small-scale design space benchmarks such as \textit{atax}, LLMMH variants maintain comparable solution quality with meta-heuristic methods. Comprehensive experimental results demonstrate that LLMACO(deepseek-r1) achieves the optimal performance with the lowest ADRS, showing improvements of 89.34\%, 85.90\%, and 88.07\% over SA, NSGA-II, and ACO respectively. While LLMMH algorithms exhibit marginally inferior performance compared to traditional metaheuristics on small-scale design space benchmark due to LLMs' inherent hallucination issues (a longstanding challenge in NLP where models generate semantically plausible but technically invalid outputs), holistic evaluation reveals the framework's superior overall performance across metrics. This demonstrates LLMMH's effectiveness in balancing exploration-exploitation tradeoffs despite occasional localized solution quality variations. This quantitatively verifies that our LLMMH framework not only resolves inherent limitations of meta-heuristics but also substantially outperforms them in solution quality. Based on the overall runtime of each algorithm on five benchmarks (see Table \ref{table 6}), we also observed that the computational overhead of LLMMH variants remains significant due to the inference latency of large language models (particularly o3-mini and deepseek-r1), compounded by per-iteration API calls and network delays.

	To further validate the effectiveness of our proposed method, we selected the top-performing combination LLMACO (deepseek-r1) based on prior DSE experiments and compared it against SOTA methods GNN-DSE \cite{Sohrabizadeh} and IRONMAN-PRO \cite{Wu2}. The evaluation comprised two components: for GNN-DSE, algorithms were executed under identical time constraints (using our method's runtime per benchmark) with results benchmarked against LLMACO (deepseek-r1); for IRONMAN-PRO, the total number of exploration points was constrained, and comparisons were conducted against its AC-FT and PG-FT algorithms. As shown in Table \ref{cl}, under the same running time, compared with the exact algorithm GNN-DSE, LLMACO (deepseek-r1) reduces the ADRS by 68.17\%. Compared with the AC-FT and PG-FT algorithms proposed in IRONMAN-PRO, LLMACO (deepseek-r1) can find better solutions within a limited number of explorations, with its ADRS reduced by 63.07\% and 59.98\% respectively.

	\begin{table*}
		\renewcommand{\arraystretch}{1.2}
		\centering
		\small
		\caption{Experimental Results of LLMACO(deepseek-r1), GNN-DSE, and IRONMAN-PRO on Unseen Kernels.}
		\label{cl}
		\begin{tabular}{l | c | c | c | c | c}
			\toprule
			{\bf{Benchmark}}&\multicolumn{2}{c|}{\bf{LLMACO(deepseek-r1)}}&\bf{GNN-DSE}&\multicolumn{2}{c}{\bf{IRONMAN-PRO}} \\
			\midrule
			~ & Runtime(s) & ADRS & ADRS & AC-FT ADRS & PG-FT ADRS \\ 
			\hline
			atax & 83 & 0.0086 & 0.1482 & 0.0662& 0.0087 \\ 
			\hline
			heat-3d & 167 & 0 & 0.0005 & 0.0357& 0.0480 \\
			\hline
			doitgen & 6 & 0 & 0 & 0.2969& 0.2994\\
			\hline
			gemm-p & 250 & 0.0096 & 0.2665 & 0.0121& 0.0370\\
			\hline
			mvt & 334 & 0.1516 & 0.1174 & 0.0477& 0.0305\\
			\hline
			Average ADRS & - & {\bf{0.0339}} & 0.1065 & 0.0918& 0.0847\\
			\hline
			\multicolumn{3}{c|}{\bf{LLMACO(deepseek-r1) Improv. over GNN-DSE}}&\multicolumn{3}{c}{\bf{68.17\%}}\\
			\multicolumn{3}{c|}{\bf{LLMACO(deepseek-r1) Improv. over AC-FT}}&\multicolumn{3}{c}{\bf{63.07\%}} \\
			\multicolumn{3}{c|}{\bf{LLMACO(deepseek-r1) Improv. over PG-FT}}&\multicolumn{3}{c}{\bf{59.98\%}} \\
			\bottomrule
		\end{tabular}
		\vspace{-3mm}
	\end{table*}

	\section{Conclusion}
	In this paper, we present the ECoGNNs-LLMMH framework, which comprises a GNN prediction model that dynamically adapts message-passing strategies based on task-specific characteristics and LLM-driven meta-heuristic algorithms for DSE. ECoGNNs enhance LLMMH's exploration capabilities by providing high-precision QoR predictions for design configurations without relying on time-consuming EDA flows. Evaluation across multiple metrics demonstrates the clear superiority of the proposed ECoGNN and LLMMH algorithm over current SOTA approaches. This work presents the first systematic implementation of LLMs in DSE. Beyond demonstrating LLMs' capability to autonomously execute complex operations traditionally requiring human expertise, we aim to address the limitations of conventional heuristic-based methodologies that dominate current DSE practices. More significantly, our LLMMH framework demonstrates superior performance compared to traditional heuristic approaches. The proposed paradigm shift of leveraging LLMs for domain-specific knowledge-intensive tasks establishes a new methodology with promising potential for widespread adoption in DSE research and applications. The current implementation of LLMMH presents valuable opportunities for enhancement in future endeavors. By addressing specific areas for improvement, we can significantly elevate its effectiveness and impact.
	\begin{itemize}
		\item [1)] {\bf{Improving the selection of LLMs}} The current DSE experiment implemented multiple LLMs, though the investigation does not comprehensively encompass all established model archetypes. Notably, the hypothesis that LLMs specifically designed for mathematical reasoning and code generation might demonstrate superior configuration generation capabilities compared to conversational LLMs in DSE contexts remains to be rigorously validated through systematic comparative analysis.
		\item [2)] {\bf{Reducing the runtime of LLMMH}} In the experiments, we employed multiple LLMs, necessitating the implementation of LangChain for online LLM orchestration. However, the algorithms within our proposed LLMMH framework currently exhibit elevated runtime overhead. This bottleneck is not insurmountable — potential optimizations include developing locally deployed distilled LLMs specialized for DSE tasks (given the correlation between LLM inference latency and parameter count). The approach holds significant promise for runtime reduction, constituting a key focus of our ongoing research efforts.
		\item [3)] {\bf{Optimizing prompt engineering techniques}} 
		Experimental results empirically demonstrate that the output quality of LLMs exhibits strong dependence on prompt configuration design. Dedicated prompt engineering techniques, such as self-reflective prompts \cite{Shinnp} and directional-stimulus prompts \cite{Lip}, may lead to more significant performance enhancements in LLMMH's evolutionary optimization processes.
	\end{itemize}
	\nocite{*}
    \bibliographystyle{unsrt}  
    \bibliography{reference}  
\end{document}